\documentclass[conference]{IEEEtran}
\IEEEoverridecommandlockouts

\usepackage{cite}
\usepackage{amsmath,amssymb,amsfonts}
\usepackage{algorithmic}
\usepackage{graphicx}
\usepackage{textcomp}
\usepackage{xcolor}

\usepackage{booktabs}
\usepackage{hyperref}
\usepackage{multirow}
\usepackage{colortbl}

\def\BibTeX{{\rm B\kern-.05em{\sc i\kern-.025em b}\kern-.08em
    T\kern-.1667em\lower.7ex\hbox{E}\kern-.125emX}}
\begin{document}

\title{Semantic Residual for Multimodal Unified Discrete Representation\\
}

\author{\IEEEauthorblockN{1\textsuperscript{st} Hai Huang}
\IEEEauthorblockA{
\textit{School of Software Technology} \\
\textit{Zhejiang University}\\
Ningbo, China \\
haihuangcode@outlook.com}
\and
\IEEEauthorblockN{2\textsuperscript{rd} Shulei Wang}
\IEEEauthorblockA{
\textit{School of Software Technology} \\
\textit{Zhejiang University}\\
Ningbo, China \\
shuleiwang@zju.edu.cn}
\and
\IEEEauthorblockN{3\textsuperscript{nd} Yan Xia{\footnotesize \textsuperscript{*}}
\thanks{*Corresponding Author}
}
\IEEEauthorblockA{
\textit{College of Computer Science and Technology} \\
\textit{Zhejiang University}\\
Hangzhou, China \\
xiayan.zju@gmail.com}


}

\maketitle


\begin{abstract}
Recent research in the domain of multimodal unified representations predominantly employs codebook as representation forms, utilizing Vector Quantization(VQ) for quantization, yet there has been insufficient exploration of other quantization representation forms. Our work explores more precise quantization methods and introduces a new framework, Semantic Residual Cross-modal Information Disentanglement (SRCID), inspired by the numerical residual concept inherent to Residual Vector Quantization (RVQ). SRCID employs semantic residual-based information disentanglement for multimodal data to better handle the inherent discrepancies between different modalities. Our method enhances the capabilities of unified multimodal representations and demonstrates exceptional performance in cross-modal generalization and cross-modal zero-shot retrieval. Its average results significantly surpass existing state-of-the-art models, as well as previous attempts with RVQ and Finite Scalar Quantization (FSQ) based on these modals. 
\end{abstract}


\section{Introduction}
Different modalities contain distinctly different information; for example, sounds present in audio may not have corresponding visual sources in video. Many researchers have addressed how to learn a unified representation from these modalities, offering various solutions~\cite{radford2021learning,luo2022clip4clip,xu2021videoclip}. Some approaches employ modality-agnostic encoders to encode different modalities~\cite{akbari2021vatt, you2022learning}, while others use contrastive learning to align representations across modalities~\cite{radford2021learning, liang2022mind}. Beyond continuous representation spaces, some studies have opted for codebooks as unified representations for better interpretability~\cite{duan2022multi,liu2021cross,lu2022unified,zhao2022towards,xia2024achieving,huang2024unlocking}, achieving notable results. However, discussions on the forms of unified representations remain superficial, with little exploration of quantization methods beyond VQ.

We begin with the most intuitive switch from VQ~\cite{van2017neural,ji2024wavtokenizer} to RVQ~\cite{lee2022autoregressive}, which has seen effective applications in many fields such as speech~\cite{ji2024controlspeech}, image~\cite{lee2022autoregressive} and retrieval~\cite{fang2024ace}, yet is scarcely utilized in the research of multimodal unified representations. We also experimente with FSQ~\cite{mentzer2023finite} that does not require a codebook, with detailed results presented in Section~\ref{sec:experiment}. The findings suggest that neither RVQ nor FSQ improved the model's performance, prompting the question: Does a deeper, more precise quantization method not contribute to multimodal unification? We hypothesize that this is because any form of quantization inherently seeks more accurate quantization of the original data, which is suitable for unimodal representation tasks but not for multimodal unified representations. Excessive focus on precise quantization of modality 'A' could compromise the precise of quantization for modality 'B'. 

\begin{figure*}[h]
\centerline{\includegraphics[width=0.8\textwidth]{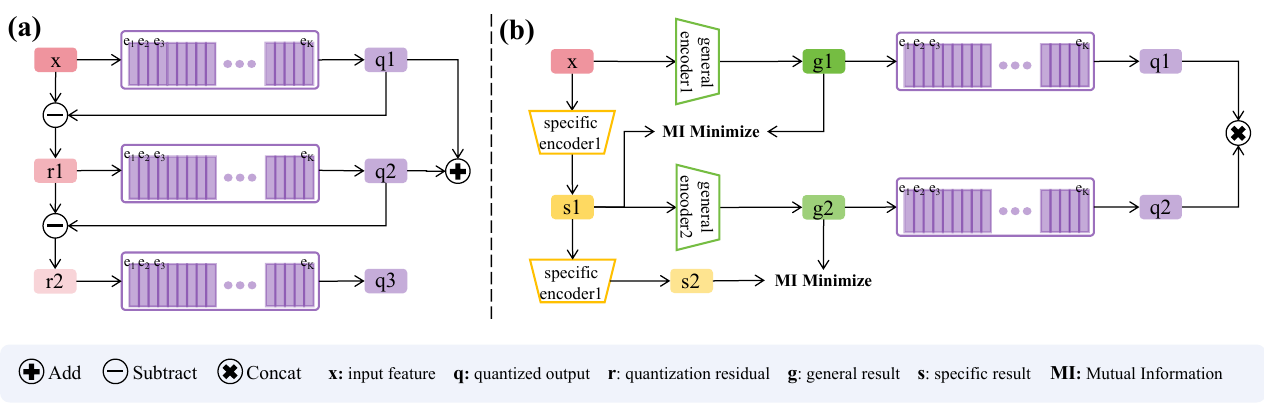}}
\caption{(a) RVQ (b) Simplified diagram of SRCID}
\label{fig:RVQandSRCID}
\vspace{-5mm}
\end{figure*}

Consequently, we believe that RVQ's approach of numerical residuals is not suitable for multimodal unified representations. We have decided to start with semantic residuals to achieve better multimodal representations. As shown in Fig.~\ref{fig:RVQandSRCID}, if we consider mutual information minimize, results of general encoder and specific encoder in subplot (b) as a form of subtraction, quantized output and quantization residual in subplot (a), it becomes apparent that the structure of SRCID and RVQ are quite similar. However, RVQ deals with numerical residuals, meaning that $q2$ lacks inherent semantics and only functions effectively when combined with $q1$. In contrast, SRCID addresses semantic residuals, extracting the remaining modal-general results $g2$ from $s1$ as a new, independent semantic entity. 


Our contributions can be summarized as follows:
\begin{itemize}
    \item We explore the impact of different quantization methods on constructing a unified representation space and demonstrate that quantization approaches focused on precision are not directly suitable for the multimodal unification domain.
    \item We introduce a new framework for multimodal unified discrete representation, named Semantic Residual Cross-modal Information Disentangling (SRCID), which, unlike the numerical residuals of RVQ, implements semantic residuals. This approach achieves results that significantly surpass those of RVQ and previous state-of-the-art (SOTA) models.
\end{itemize}

\section{Method}
\label{SRCID}
\begin{figure}[h]
\centerline{\includegraphics[width=0.5\textwidth]{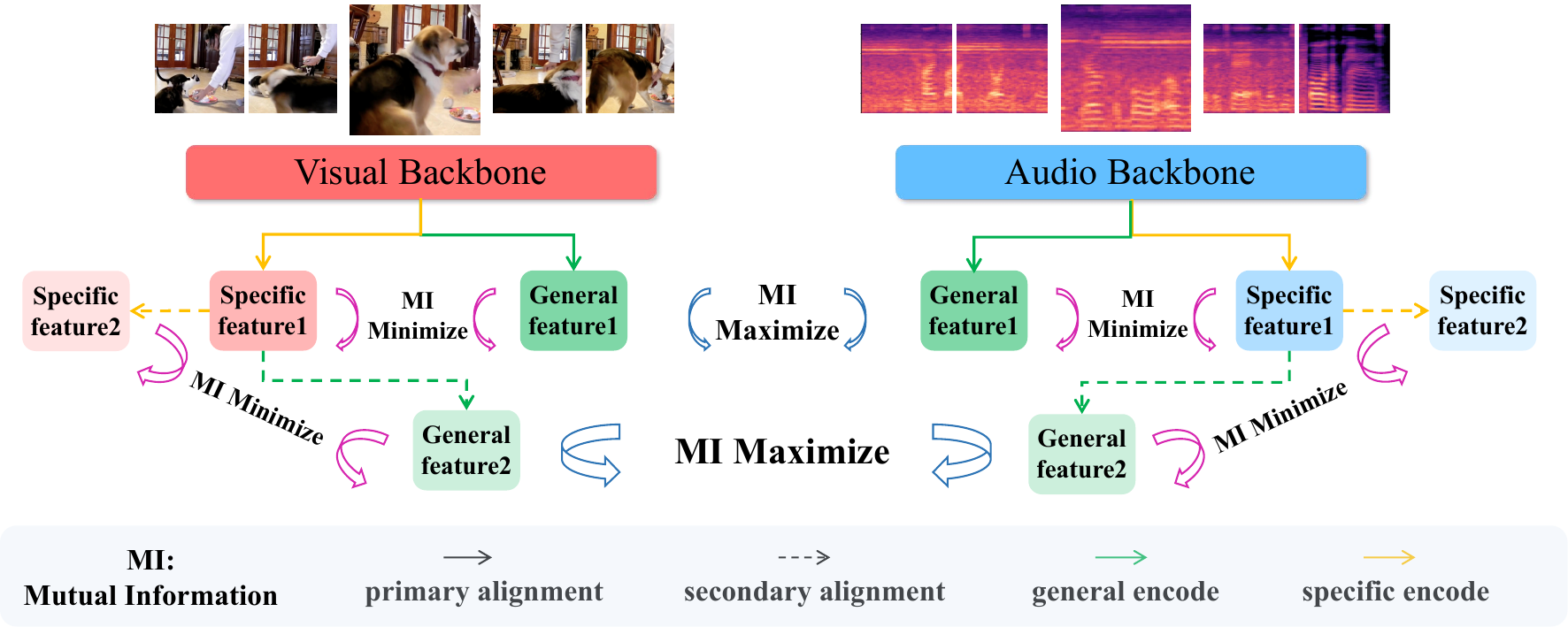}}
\caption{SRCID Encoder Framework}
\label{fig:SRCID}
\vspace{-5mm}
\end{figure}

As shown in Fig.~\ref{fig:SRCID}, SRCID architecture is divided into two layers. The first layer utilizes mutual information to separate the primary modal-general results from modal-specific results. Building on the converged training of the first layer, the second layer further disentangles the modal-specific results obtained from the first layer, thus achieving semantic residuals. The core components of both layers involve minimizing mutual information between modal-general and modal-specific results within each modality using CLUB~\cite{cheng2020club}, and maximizing mutual information among modal-general results across different modalities using Cross-Modal CPC~\cite{oord2018representation}.

Given three paired modalities, ${(\mathbf{x}^{a}_{i},\mathbf{x}^{b}_{i},\mathbf{x}^{c}_{i})}^{N}_{i=1}$, we employ three modal-specific encoders $\Psi^{a},\Psi^{b},\Psi^{c}$ to extract modal-specific features $\overline{\mathbf{z}}^{a}_{i},\overline{\mathbf{z}}^{b}_{i},\overline{\mathbf{z}}^{c}_{i}$, and three modal-general encoders $\Phi^{a},\Phi^{b},\Phi^{c}$ to extract modal-general features $\mathbf{z}^{a}_{i}, \mathbf{z}^{b}_{i}, \mathbf{z}^{c}_{i} \in \mathbb{R}^{T \times D}$ from modalities A, B, and C, respectively. To enhance the effectiveness of multimodal representations, SRCID extends this process to two layers, $m \in \{a,b,c\}, \ k \in \{1,2\}$:
\begin{equation}
\begin{split}
    \mathbf{z}^{m}_{i,k} &= \Phi^{m}_{k}(\mathbf{x}^{m}_{i,k}), 
    \overline{\mathbf{z}}^{m}_{i,k} = \Psi^{m}_{k}(\mathbf{x}^{m}_{i,k}), 
\end{split}
\label{equ1} 
\end{equation}

Furthermore, the input to all second-layer encoders is the output of the first-layer modal-specific results:
\begin{equation}
    \mathbf{x}^{m}_{i,2} = \overline{\mathbf{z}}^{m}_{i,1}.
\label{equ1} 
\end{equation}

The latent codebook $\mathbf{e} \in R^{L\times D \times K}$ is shared across modalities A, B, and C, where T, L, D, and K represent time, size of the discrete latent space, hidden dimension, and layer num of codebook, respectively. The dimension of $\overline{\mathbf{z}}^{a}_{i,k},\overline{\mathbf{z}}^{b}_{i,k},\overline{\mathbf{z}}^{c}_{i,k}$ various for different modalities. Apply vector quantized operation to map model-general feature $\mathbf{z}^{a}_{i,k}, \mathbf{z}^{b}_{i,k}, \mathbf{z}^{c}_{i,k}$ to discrete latent codes, $t \in [0, T)$:
\begin{equation}
\begin{split}
    \hat{\mathbf{z}}^{m}_{i,k,t} &= VQ(\Phi^{m}_{k}(\mathbf{x}^{m}_{i,k})) = VQ(\mathbf{z}^{m}_{i,k,t}) = e_{l,k}, \\
     {\rm where} \ l &= argmin_{j}\lvert\lvert\Phi_{k}(x) - e_{j,k}\rvert\rvert_{2}
\label{equ2} 
\end{split}
\end{equation}

Then, we combine $\mathbf{\hat{z}}_{i,k}^{m}$ with $\mathbf{\bar{z}}_{i,k}^{m}$ together to reconstruct original features:

\vspace{-4mm}
\begin{equation}
\begin{split}
    \sum_{k=1}^K \left( \underbrace{\|\mathbf{x}_{i,k}^{m} - D(\hat{\mathbf{z}}_{i,k}^{m};\bar{\mathbf{z}}_{i,k}^{m})\|_2^2}_{\text{reconstruction loss}} \right. & + \underbrace{\|\operatorname{sg}[\phi^{m}_{k}(\mathbf{x}_{i,k}^{m})] - \mathbf{e}\|_2^2}_{\text{VQ loss}} \\
    & \left. + \underbrace{\beta \|\phi^{m}_{k}(\mathbf{x}_{i,k}^{m}) - \operatorname{sg}[\mathbf{e}]\|_2^2}_{\text{commitment loss}} \right)
\end{split}
\label{equ3}
\end{equation}

where $\operatorname{sg}$ denotes the stop gradient operation, and $\beta$ is set to 0.25. We employ the Exponential Moving Average (EMA) strategy to replace the VQ loss, and utilize MMEMA\cite{xia2024achieving} to modify the commitment loss. The reconstruction loss ensures that the compressed latent codes $e_{l}$ retain the general information of different modalities. Ideally, $\mathbf{z}_{i,k}^{a}$, $\mathbf{z}_{i,k}^{b}$, and $\mathbf{z}_{i,k}^{c}$, encoded from different modalities with the same semantics, should be mapped to the same discrete latent code. However, in the absence of effective supervision, the presence of a modality gap may lead to $\mathbf{z}_{i,k}^{a}$, $\mathbf{z}_{i,k}^{b}$, and $\mathbf{z}_{i,k}^{c}$ converging to distinct regions of the codebook\cite{zhao2022towards, liu2021cross}. Therefore, we aim to minimize the mutual information between the general and specific results within a single modality and maximize the mutual information across the general results of different modalities.

{\textbf{Mutual Information Minimization: }} CLUB~\cite{cheng2020club} could optimize the mutual information upper bound, demonstrating superior advantages in information disentanglement. Given two variables $\mathbf{x}$ and $\mathbf{y}$, the objective function of CLUB is defined as:
\begin{equation}
\begin{split}
    I_{vCLUB}(\mathbf{x};\mathbf{y})&:=\mathbb{E}_{p(\mathbf{x},\mathbf{y}}[\log q_{\theta}(\mathbf{y}|\mathbf{x})] 
    \\&-\mathbb{E}_{p(\mathbf{x})}\mathbb{E}_{p(\mathbf{y})}[\log q_{\theta}(\mathbf{y}|\mathbf{x})],
\label{equ4} 
\end{split}
\end{equation}

where $q_{\theta}$ is the variational approximation of ground-truth posterior of $\mathbf{y}$ given $\mathbf{x}$ and can be parameterized by a network $\theta$. We use CLUB to optimize the MI upper bound between the modal-general features $\mathbf{z}^{m}_{i,k}$ and modal-specific features $\overline{\mathbf{z}}^{m}_{i,k}$:

\vspace{-4mm}
\begin{equation}
\begin{split}
    \hat{I}_{vCLUB}&=\frac{1}{N}\frac{1}{K}\sum_{i=1}^N\sum_{k=1}^K[\frac{1}{T}\sum_{t=1}^T\log q_{\theta}(\overline{\mathbf{z}}^{m}_{i,k}|\mathbf{z}^{m}_{i,k})\\&- \frac{1}{N}\frac{1}{T}\sum_{j=1}^N\sum_{t=1}^T\log q_{\theta}(\overline{\mathbf{z}}^{m}_{j,k}|\mathbf{z}^{m}_{i,k})]
\label{equ5} 
\end{split}
\end{equation}

\textbf{Mutual Information Maximization: }Contrastive Predictive Coding (CPC)\cite{oord2018representation} aims to maximize the mutual information between sequence items by predicting future samples using autoregressive models and is widely adopted in self-supervised learning, We utilize it to facilitate the alignment of modal-general information. Given the general features $\mathbf{z}^{a}, \mathbf{z}^{b}, \mathbf{z}^{c} \in \mathbb{R}^{T\times D}$, a prediction horizon of R steps, and a random time moment $t \in (0, \text{T-R}]$, two single-layer unidirectional LSTMs are used to summarize the information of all $\mathbf{z}^{a}_{\leq t}, \mathbf{z}^{b}_{\leq t}, \mathbf{z}^{c}_{\leq t}$, yielding three context representations as $\mathbf{c}^{m}_{t}$ = LSTM($\mathbf{z}^{m}_{\leq t} \in \mathbb{R}^{D}, m \in {a,b,c}$).

For modality M, we first select a set $Z^{n}$ of N-1 random negative samples and one positive sample $\mathbf{z}^{n}_{t+r}$ from modality N, then use $\mathbf{c}^{m}_{t}$ to predict r-th future step $\mathbf{z}^{n}_{t+r}$ in modality N, and the InfoNCE loss for all modality can be optimized as:

\begin{equation}
\begin{split}
    L^{m2n}_{cpc} = -\frac{1}{R}\frac{1}{K}\sum_{r=1}^R\sum_{k=1}^K\log[\frac{\exp{(\mathbf{z}^{n}_{t+r,k}W^{m}_{r,k}\mathbf{c}^{m}_{t,k})}}{\sum_{\mathbf{z}_{j}\in Z_{n}}\exp{(\mathbf{z}^{n}_{j,k}W^{m}_{r,k}\mathbf{c}^{m}_{t,k})}}],
\label{equ6} 
\end{split}
\end{equation}

The overall objective of SRCID is a combination of these loss functions across both layers:

\begin{equation}
    L = L_{\text{recon}} + L_{\text{commit}} + L_{\text{cpc}} + L_{\text{cmcm}} + L_{\text{MI}},
\end{equation}
where $L_{\text{recon}}$ is the reconstruction loss that merges the modal-specific and modal-general results for each modality and compares them with the original input using MSE loss, $L_{\text{commit}}$ is the commitment loss that computes the MSE loss between the modal-general results and their quantized codes, $L_{\text{cpc}}$ is the Contrastive Predictive Coding loss that enhances cross-modal alignment and inference by predicting future samples in one modality using information from another, $L_{\text{cmcm}}$ is the objective loss proposed in \cite{liu2021cross}, which also promotes the alignment among modalities, and $L_{\text{MI}} = \hat{I}_{vCLUB}$ represents the mutual information loss concerning the modal-specific and modal-general results within each modality. The VQ loss is replaced by MMEMA, so it does not appear in the final loss.

\section{Experiment}
\label{sec:experiment}
\subsection{Datasets and Tasks}

\textbf{Pre-train: }The pretraining dataset uses the VGGsound-AVEL 40k~\cite{chen2020vggsound,zhou2022contrastive} with prompts provided by~\cite{xia2024achieving}.

\textbf{Downstream: }The unified representation pre-trained models will be evaluated on several downstream tasks using different datasets. {\bf Cross-modal event classification on AVE dataset:}~\cite{avel} training on one modality (video) and evaluating on another (audio). {\bf Cross-modal event localization on AVVP dataset:}~\cite{tian2020unified} localizing events in one modality and transferring to the other. {\bf Cross-dataset localization/classification:} training on classification in AVE and evaluating localization in AVVP, transferring across datasets. Cross-modal classification between UCF-101~\cite{soomro2012ucf101} visual clips and VGGSound-AVEL audio clips.  {\bf Cross-modal Zero-shot Retrieval:} MSCOCO~\cite{lin2014microsoft}; we adopt a process similar to the test set~\cite{test1k} consists of 1000 video-text pairs from MSRVTT~\cite{msrvtt} to get 500 pairs. This task tests the retrieval effectiveness between videos and text. Clotho~\cite{drossos2020clotho}; assesses the zero-shot retrieval capability for audio-text alignment.


\subsection{Implementation Details}
We compare our method with recent unified representation approaches including DCID~\cite{xia2024achieving}, CODISS~\cite{duan2022multi}, TURN~\cite{zhao2022towards}, and CMCM~\cite{liu2021cross}. These methods are implemented across our tasks, with their effectiveness assessed on four downstream tasks involving datasets such as AVE~\cite{avel}, VGGSound-AVEL~\cite{zhou2022contrastive, zhou2021positive}, and UCF101~\cite{soomro2012ucf101}, where precision is the evaluation metric. For the AVVP dataset~\cite{tian2020unified}, accuracy serves as the measure, while recall is utilized for evaluating cross-modal zero-shot retrieval.

The losses of SRCID are not suitable for full backpropagation initially. If subsequent layer losses participate in backpropagation before sufficient disentanglement is achieved by previous layers, this can lead to failure in aligning modal-general semantics from coarse to fine granularity across layers. We use a warm-start technique, applying only the preceding layer's loss during the initial epochs. The second layer's loss is activated once the first layer's MI loss approaches 0, signaling successful disentanglement. Only then are losses from subsequent layers introduced into the backpropagation.
All results presented in Table~\ref{tab:quantization}, \ref{tab:cross-modal generalization}, \ref{tab:zero-shot retrieval}  were obtained with a codebook size set to 400 and an embedding dimension set to 256. The backbone models used to extract features for video, audio, and text modalities are VGG19~\cite{simonyan2014very}, VGGish~\cite{hershey2017cnn}, and BERT~\cite{devlin2018bert}, respectively.

\subsection{Performance Analysis}
\label{sec:preformance_analysis}

\begin{table*}[h]
\caption{Impact of different quantization methods on DCID~\cite{xia2024achieving}: grey rows indicate results for the training modality of the previous row and their averages exclude AVE$\rightarrow$AVVP, while averages for the remaining rows include eight results.(AVE$\rightarrow$AVVP's $m\rightarrow m, m\in \{A,V\}$ same as AVE)}
\centering
\resizebox{0.7\textwidth}{!}
{
\begin{tabular}{cccccccccc}
\toprule
\multirow{1}{*}{Method}                                                                                           
                                 &
                                 \multicolumn{2}{c}{\begin{tabular}[c]{@{}c@{}}AVE\\ V$\rightarrow$A   A$\rightarrow$V\end{tabular}} & 
                                 \multicolumn{2}{c}{\begin{tabular}[c]{@{}c@{}}AVVP\\ V$\rightarrow$A   A$\rightarrow$V\end{tabular}} & 
                                 \multicolumn{2}{c}{\begin{tabular}[c]{@{}c@{}}AVE$\rightarrow$AVVP\\ V$\rightarrow$A   A$\rightarrow$V\end{tabular}} & 
                                 \multicolumn{2}{c}{\begin{tabular}[c]{@{}c@{}}UCF(v)$\leftrightarrow$VGG(a)\\ V$\rightarrow$A   A$\rightarrow$V\end{tabular}}&
                                 \multicolumn{1}{c}{\begin{tabular}[c]{@{}c@{}}Avg.\\ \end{tabular}}\\
                                 \hline
DCID(VQ)    &54.1 & 55.0 & 63.4  & 71.0  & 53.0 & 52.4 & 67.1 & 60.6 & 59.58\\ 
\rowcolor{gray!20} 
{$m\rightarrow m$}   & {64.8} & {65.8} & {71.0}  & {72.9}  & {-}  & {-} &80.0    & 85.4 & 73.32\\
DCID(FSQ)&45.2 & 50.4 & 48.1  & 54.9  & 51.4 & 44.0 & 67.1    & 60.6 & 52.71\\ 
\rowcolor{gray!20} 
{$m\rightarrow m$}& {76.5} & {78.2} & {65.8}  & {72.1}  & {-}  & {-} &98.5    & 92.2 & 80.55\\
DCID(RVQ-2)& 49.3 & 55.0 & 58.4  & 66.9  & 57.4 & 54.2 & 68.9    & 63.5 & 59.20\\ 
DCID(RVQ-3)& 47.6 & 53.2 & 59.9  & 67.2  & 54.3 & 53.0 & 68.5    & 62.4 & 58.26\\ 
DCID(RVQ-4)& 48.5 & 55.5 & 58.7  & 68.6  & 56.3 & 54.7 & 69.4    & 64.3 & 59.50\\ 
\rowcolor{gray!20} 
{$m\rightarrow m$}   & {69.9} & {70.3} & {75.1}  & {75.6}  & {-}  & {-} &88.0    & 90.8 & 78.28\\
\bottomrule
\end{tabular}
}
\label{tab:quantization}
\vspace{-1mm}
\end{table*}

\begin{table*}[h]
\caption{Comparison with SOTA methods on four audiovisual downstream tasks.("only1" refers to using only the first layer to obtain modal-general results during the inference phase of SRCID)}
\centering
\resizebox{0.7\textwidth}{!}{%
\begin{tabular}{cccccccccc}
\toprule
\multirow{1}{*}{Method}                                                                                         
                                 &
                                 \multicolumn{2}{c}{\begin{tabular}[c]{@{}c@{}}AVE\\ V$\rightarrow$A   A$\rightarrow$V\end{tabular}} & 
                                 \multicolumn{2}{c}{\begin{tabular}[c]{@{}c@{}}AVVP\\ V$\rightarrow$A   A$\rightarrow$V\end{tabular}} & 
                                 \multicolumn{2}{c}{\begin{tabular}[c]{@{}c@{}}AVE$\rightarrow$AVVP\\ V$\rightarrow$A   A$\rightarrow$V\end{tabular}} & 
                                 \multicolumn{2}{c}{\begin{tabular}[c]{@{}c@{}}UCF(v)$\leftrightarrow$VGG(a)\\ V$\rightarrow$A   A$\rightarrow$V\end{tabular}}&
                                 \multicolumn{1}{c}{\begin{tabular}[c]{@{}c@{}}Avg.\\ \end{tabular}}\\
                                 \hline
CODIS\cite{duan2022multi}   &36.8 & 39.7 & 46.7  & 47.6  & 37.8 & 37.6 & 50.8 & 45.2 & 42.78\\ 
TURN\cite{zhao2022towards}    &37.6 & 39.2 & 48.4  & 50.2  & 36.6 & 38.4 & 49.4 & 46.1 & 43.24\\ 
CMCM\cite{liu2021cross}    &46.3 & 45.8 & 57.1  & 58.2  & 44.1 & 45.2 & 51.2 & 48.3 & 49.53\\ 
DCID\cite{xia2024achieving}    &\textbf{54.1} & \textbf{55.0} & 63.4  & 71.0  & 53.0 & 52.4 & 67.1 & 60.6 & 59.58\\ 
SRCID(only1)& 49.4 & 53.5 & 60.4  & 77.2  & 53.4 & 52.9 & \textbf{70.6}    & \textbf{59.5} & 59.61\\ 
SRCID & 49.2 & 54.2 & \textbf{72.1}  & \textbf{85.6}  & \textbf{54.8} &\textbf{53.7} & 69.1    & 59.0 & \textbf{62.21}\\ 
\bottomrule
\end{tabular}%
}
\label{tab:cross-modal generalization}
\vspace{-3mm}
\end{table*}

\begin{table}[h]
\caption{Comparison with SOTA methods on three cross-modal zero-shot retrieval tasks, all results are calculated as the mean across two directions.}
\centering
\resizebox{0.45\textwidth}{!}{%
\begin{tabular}{cccccccc}
\toprule
\centering
\multirow{2}{*}{Method} & \multicolumn{3}{c}{MSCOCO(V$\leftrightarrow$T)} &  \multicolumn{3}{c}{Clotho(A$\leftrightarrow$T)} & \multirow{2}{*}{Avg.}
\\
 & R@1 & R@5 & R@10 &  R@1 & R@5 & R@10  & \\
\toprule
CMCM\cite{liu2021cross} & 0.4 & 3.8 & 6.9 &  1.62 & 8.04 & 14.87 &  5.94  \\
DCID\cite{xia2024achieving} & 0.5 & 4.5 & 7.8 &  2.06 & 9.00 & 16.70 & 6.76  \\
SRCID & \textbf{0.9} & \textbf{4.7} & \textbf{8.3} &  \textbf{2.28} & \textbf{9.54} & \textbf{17.32} &  \textbf{7.17}  \\
\bottomrule
\end{tabular}
\label{tab:zero-shot retrieval}
}
\vspace{-5mm}
\end{table}

\textbf{Quantization:} As shown in Table~\ref{tab:quantization}, we experimented with different quantization methods on the SOTA discrete unified representation model, DCID~\cite{xia2024achieving}, and found that for cross-modal tasks, the simplest method, vq, performed the best. Concurrently, observing the $m\rightarrow m$ results, we noted that FSQ and RVQ significantly outperformed VQ, confirming our hypothesis: improvements in quantization precision are specific to individual features and do not simultaneously enhance multimodal unified results. The closer the representation gets to modality 'A', the further it moves from modality 'B', leading to improvements in intramodal results but no enhancement or even a decline in cross-modal tasks.

\textbf{Cross-Modal Generalization and Zero-shot Retrieval:} We compare our model with leading methods on four downstream tasks. All models are pre-trained on same dataset sizes. It is worth mentioning that in localization tasks (avvp, ave$\rightarrow$avvp), each second of every audio-video can belong to multiple categories. In contrast, in the classification tasks discussed in this paper (ave, ucf(v)$\rightarrow$vgg(a)), the entire duration of each audio or video belongs to only one category. Therefore, localization tasks require finer granularity.

As shown in Table~\ref{tab:cross-modal generalization}, using only the first-layer modal-general results, SRCID achieves performance close to the SOTA across four tasks. When both layers of modal-general results are employed, SRCID clearly surpasses the SOTA. A closer examination of the last two rows reveals that adding the second-layer modal-general results slightly reduces performance in classification-related tasks (ave, ucf(v)$\rightarrow$vgg(a)), while significantly enhancing outcomes in localization-related tasks (avvp, ave$\rightarrow$avvp). This indicates that the results extracted from the second layer are crucial for impacting fine-grained details. Similarly, SRCID's performance on the AVE task is not optimal, mirroring results with RVQ. This suggests that there is room for improvement in our semantic residuals, which are not yet effectively separated by mutual information, unlike numerical subtraction which allows for complete disengagement.

As shown in Table~\ref{tab:zero-shot retrieval}, we conducted zero-shot retrieval experiments for Video$\leftrightarrow$Text and Audio$\leftrightarrow$Text on the MSCOCO~\cite{lin2014microsoft} and Clotho~\cite{drossos2020clotho} datasets, respectively. The results demonstrate that SRCID continues to maintain its superiority.

\begin{figure}[ht]
\centering
\includegraphics[width=0.45\textwidth]{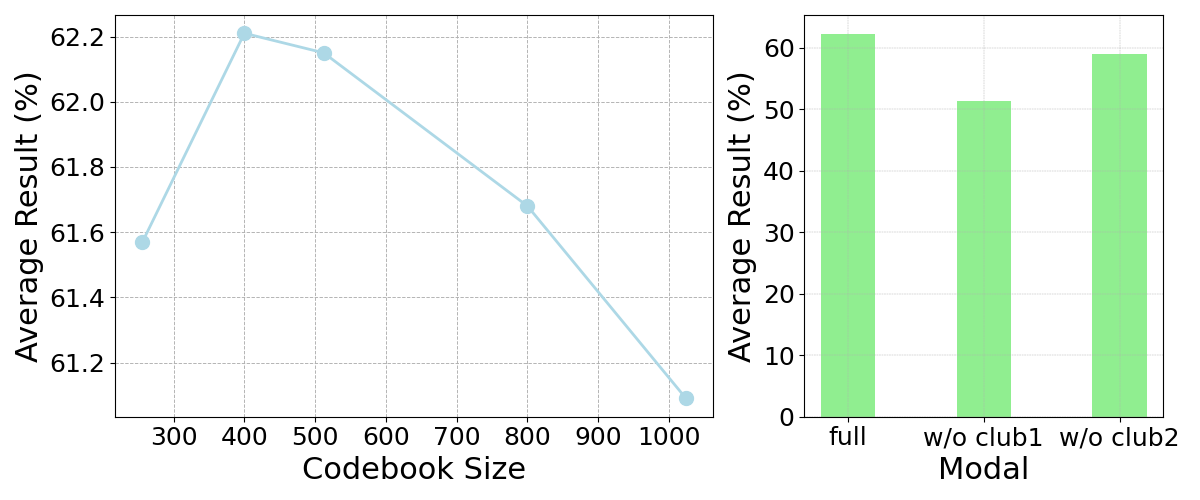}
\caption{Ablation of codebook size and club}
\label{fig:ablation}
\vspace{-5mm}
\end{figure}
\textbf{Ablation Study:} The two most critical losses in SRCID are the semantic disentanglements performed by the two layers of CLUB~\cite{cheng2020club}, as contrastive learning is almost essential for multimodal unified representations, and omitting it would undoubtedly result in a significant drop in model performance. Other losses have been proven effective in previous studies and will not be elaborated further here. Consequently, we conducted ablation experiments focusing on codebook size and CLUB effectiveness. As illustrated in Fig.~\ref{fig:ablation}, the model performs optimally with a codebook size of 400, displaying an initial increase followed by a decrease in performance. Both layers of CLUB contribute effectively, with the first layer having a more pronounced impact, as it contains the primary information.

\vspace{-2mm}
\section{Conclusion}
\label{sec:conclusion}
Previous works on multimodal unified discrete representations have not delved deeply into the forms of representation. Starting with RVQ and FSQ, we test more precise quantization methods to assess their applicability in multimodal unified representations and find that, although they perform better than VQ in unimodal scenarios, they do not aid in multimodal representation. Furthermore, inspired by the numerical residual concept of RVQ, we develop SRCID, a new framework that effectively utilizes mutual information to enhance the model's representational capabilities.


\newpage
\bibliographystyle{ieeetr}
\bibliography{ref.bib}

\end{document}